\documentclass[journal]{elsarticle}
\usepackage{amssymb}%
\usepackage{graphicx,times,amsmath,multirow,booktabs}%
\usepackage{algorithmic}%
\usepackage{algorithm}%
\usepackage{multicol}%
\usepackage{enumerate}%
\usepackage{hyperref}%
\usepackage{color}%

\begin{document}
%
%
\begin{frontmatter}%

\title{\mbox{Convex Hull-Based Multi-objective} \mbox{Genetic Programming for Maximizing ROC Performance}}%
\author[1]{Pu~Wang}%
\author[2]{Michael~Emmerich}%
\cortext[cor]{Emails: wuyou308@mail.ustc.edu.cn, emmerich@liacs.nl}%
\author[2]{Rui~Li}%
\author[1]{Ke~Tang}%
\author[2]{Thomas~B$\ddot{a}$ck}
\author[3]{Xin~Yao}%
\address[1]{%
Nature Inspired Computation and Applications Laboratory (NICAL)\\%
School of Computer Science and Technology\\%
University of Science and Technology of China (USTC), %
Hefei, Anhui 230027, China.}%
\address[2]{%
Leiden Institute for Advanced Computer Science (LIACS),\\ 
Leiden University, 2333 CA Leiden, Netherlands.}%
\address[3]{%
Center of Excellence for Research in Computational Intelligence and Applications (CERCIA)\\%
School of Computer Science, \\%
The University of Birmingham Edgbaston, \\%
Birmingham B15 2TT, U.K.}%

\begin{abstract}
Receiver operating characteristic (ROC) is usually  used to analyse the performance of classifiers in data mining. An important ROC analysis topic is ROC convex hull(ROCCH), which is the least convex majorant (LCM) of the empirical ROC curve, and covers potential optima for the given set of classifiers. Generally, ROC performance maximization could be considered to maximize the ROCCH, which also means to maximize the true positive rate (\emph{tpr}) and minimize the false positive rate (\emph{fpr}) for each classifier in the ROC space. However, \emph{tpr} and \emph{fpr} are conflicting with each other in the ROCCH optimization process. Though ROCCH maximization problem seems like a multi-objective optimization problem (MOP), the special characters make it different from traditional MOP. In this work, we will discuss the difference between them and propose convex hull-based multi-objective genetic programming (CH-MOGP) to solve ROCCH maximization problems. Convex hull-based sort is an indicator based selection scheme that aims to maximize the area under convex hull, which serves as an unary indicator for the performance of a set of points. A selection procedure is described that can be efficiently implemented and follows similar design principles than classical hypervolume based optimization algorthms. It is hypothesized that by using a tailored indicator-based selection scheme CH-MOGP gets more efficient for ROC convex hull approximation than algorithms which compute all Pareto optimal points.  To test our hypothesis we compare the new CH-MOGP to MOGP with classical selection schemes, including Non-dominated Sorting Genetic Algorithm-II (NSGA-II), Multi-objective Evolutionary Algorithms Based on Decomposition (MOEA/D) and Multi-objective Selection Based on Dominated Hypervolume (SMS-EMOA). Experimental results based on 22 well-known UCI data sets show that CH-MOGP outperforms significantly traditional EMOAs.

\end{abstract}

\end{frontmatter}%


\section{Introduction}
Traditionally, a classification task is to assign items (instances) in a data-set to target categories (classes) based on classifier(s) learnt by training instances. In binary classification there are only two classes or categories and all instances in the data set will be assigned one of them. The target of a classification problem is trying to design classifiers which make error-free assignments.

The ROC graph is a technique for visualizing, organizing and selecting classifiers based on their performance~\cite{fawcett2006introduction}. A salient topic in ROC analysis is to generate ROC curves for varying discriminative thresholds over the output of the classifier~\cite{fawcett2006introduction}, and ROC curves have been used widely in many areas. Actually, over the course of the past 40 years, ROC technique has been widely applied in many research and application areas, such as signal detection~\cite{egan1975signal}, medical decision making~\cite{sox1988medical},  diagnostic systems~\cite{swets1988measuring}.

Though ROC curve works well in many cases, recently attention of the research is also drawn towards another perspective of ROC analysis, namely ROC convex hull (ROCCH). ROCCH pays more attention to the convex hull of a set of points (hard classifiers) obtained either from sever curves (i.e., soft classifiers) or itself (hard classifier). A classifier is potentially optimal, if and only if it is a component of ROCCH, in other words, ROCCH could provide better choices than a single ROC curve to specific environments. The significance of ROCCH in ROC analysis is that for test data sets with different skewed class distributions or misclassification costs, it is always possible to choose suitable classifiers by iso-performance lines\footnote{All classifiers corresponding to the points on one line have the same expected costs.} which is translated by operating conditions of classifiers and used to identify a portion of the ROCCH~\cite{provost2001robust}. Consequently, ROCCH is emphasized in this paper and we will focus on searching a group of independently hard classifiers to maximize the ROCCH performance rather try to maximize the area under the ROC curve (AUC) of a single soft classifier.

Essentially, ROCCH is the collection of all potentially optimal classifiers in a given set of classifiers, so ROCCH maximization is to find a group of classifiers with their performance approximating the top and the left axes as near as possible in ROC space. However, ROCCH maximization is not an easy task, there are not many works focusing on how to maximize the ROCCH though it is a really important topic in classification problems. Generally, the exist works could be reviewed into two categories, ROC geometric analysis based machine learning methods and multi-objective optimization strategies based evolutionary computation methods for ROCCH maximization.

Fawcett et al.~\cite{fawcett2001using} employed C4.5 and Rule Learning (RL) systems to induce decision rules in ROC space and its advanced version PRIE was introduced in~\cite{fawcett2008prie}. It was a straight way to analysis the geometrical properties to generate decision rules to maximize the ROC performance. However, the procedure easily gets trapped in local optima.

The concavity problem in ROC analysis was researched by Flach et al.~\cite{flach2003repairing} who demonstrated how to detect and repair concavities in ROC curves. The basic idea of that work is that if a point in the concavity can be mirrored to a better point which could perform well beyond the original ROC curve. But it is not a general method to maximize the ROC performance.

ROCCER was introduced by Prati et al. in~\cite{prati2005roccer}. It was argued that ROCCER is less dependent on the previously induced rules compared with set covering algorithms to construct rule sets that have a convex hull in ROC space. However, it adopted an association rule learner to generate new rules to cover the instance space as full as possible. It is too easy to fall into overfitting, because it needs many rules to cover the space which is similar with a decision tree with a very high height.

The Neyman-Pearson lemma as the theoretical basis for finding the optimal combination of classifiers to maximize the ROCCH is given in~\cite{Barreno_Cardenas_Tygar_2008}. In contrast to the similar technique in~\cite{flach2003repairing}, it not only focuses on repairing but it also pays attention on improving if there was on concavity. For a given rule set, the method proposed by~\cite{Barreno_Cardenas_Tygar_2008} can be efficient to combine these rules using \emph{AND} and \emph{OR} to get the optimum rule subset. However, as mentioned above, it misses schemes for generating new rules in the global rule set searching.

To maximize ROCCH is searching a group of classifiers to maximize the ROCCH performance ideally would yield classifiers that simultaneously minimize the \emph{fpr} and maximize the \emph{tpr}, i.\,e. that are located as much to the left and to the top of the ROC space as possible.  However, it is very hard to optimize \emph{fpr} and \emph{tpr} simultaneously because they are conflicting targets. From this perspective, ROCCH maximization problem is similar to multi-objective optimization problem.

Zhao~\cite{zhao2007multi} proposed specific non-dominated relationship involved into multi-objective optimization framework to optimize \emph{tpr} and $1-$\emph{fpr}. However, it paid more attention on cost-sensitive classification and made many rules by information of costs of misclassification to rank the individuals in its multi-objective genetic programming. First, it is not a general method for ROCCH maximization because it only focused on cost-sensitive problem. Second, two data sets involved in experiments are too few to evaluation the proposed method.

Bhowan et al. searched the Pareto front to maximize the accuracy of each minority class with unbalanced data set~\cite{bhowan2009multi}, and they also employed multi-objective optimization techniques to evolve diverse ensembles using genetic programming to maximize the classification performance in~\cite{bhowan2012evolving}.

Wang et al. investigated  investigated some EMOAs such as NSGA-II~\cite{deb2002fast}, MOEA/D~\cite{zhang2007moea}, SMS-EMOA~\cite{Beume20071653} and Approximation-Guided Evolutionary Multi-objective Algorithm (AG-EMOA)~\cite{Bringmann}. These different evolutionary multi-objective optimization frameworks had been combined with genetic programming to maximize ROC performance~\cite{wang2012multiobjective}.

However, ROCCH is different from Pareto front though it was reported they were similar to each other~\cite{fawcett2004roc}. ROCCH is the collection of points which construct the convex hull of existing classifiers in ROC space, and Pareto front is the collection of points that is the first level sorted by dominance relationship. Though evolutionary multi-objective algorithms(EMOAs) have been successfully used into ROCCH maximization, these EMO techniques do not take into account a special characteristic of ROCCH. That is by mixing two classifiers we can take any two real classifiers to construct any virtual classifier with its performance as a point along the line connected by above two points~\cite{fawcett2004roc}. Consequently, hard classifiers in concave parts of the Pareto front can always be replaced by classifier combinations that yield dominating points. The computational resources for the approximation of concave parts are thus better spent on the accurate approximation of only those parts of the Pareto front that are part of the convex hull.

In~\cite{shan2009multi,DavoodiMonfared20111435,ZapotecasMartinez:2010}, convex hull concept of was employed into EMOAs to make the sort fast or maintain a well-distributed set of non-dominated solutions. These work are good to supply some ideas of convex hull based sort. In~\cite{CococcioniCHEA} and~\cite{ducange2010multi}, convex hull-based ranking involved with evolutionary multi-objective optimization and fuzzy rule-based binary classifiers to maximize ROOCH in ROC space. However, the number of levels was pre-defined as three without explaining in first work and the second one was considered as bi-objectives optimization, which were accuracy of classification and complexity of classifier rules.

Moreover, instead of designing algorithms based on Pareto dominance compliant performance indicators, such as the hypervolume indicator as done in \cite{Beume20071653} and in \cite{igel2007covariance}, it seems more promising to directly target the algorithm towards the maximization of the area under the convex hull (AUCH).

In this paper, we utilize Genetic Programming (GP) combined multi-objective techniques to get the optimal ROCHH. Two strategies will be represented, the first is the convex hull-based without redundancy sort to make the population of GP into several levels such as non-dominated sort in NSGA-II, the second is using area-based contribution to select the survivors in the same level, actually we use $\mu$ + $\mu$ selection strategy as~\cite{igel2007covariance}. We show that convex hull-based without redundancy sort plays a key role in multi-objective genetic programming (MOGP) for maximizing ROCCH performance and area-based contribution selection scheme also can improve the performance.

This paper is organized as follows: Section~\ref{section:rochhmo} will discuss the relationship between ROCCH optimization and traditional multi-objective optimization in detail. Convex hull-based multi-objective genetic programming (CH-MOGP) will be described in Section~\ref{section:chmogp}. Experiments are studied in Section~\ref{section:experiment} and shows the advantages of our new algorithm. Section~\ref{section:confusion} gives the conclusions and a discussion on the important aspects and the future perspectives of this work.

\section{ROC Convex Hull and Multi-objective Optimization}
\label{section:rochhmo}

\subsection{What is ROCCH?}
\begin{figure}[htbp]
\centering
\includegraphics[width=3.5in]{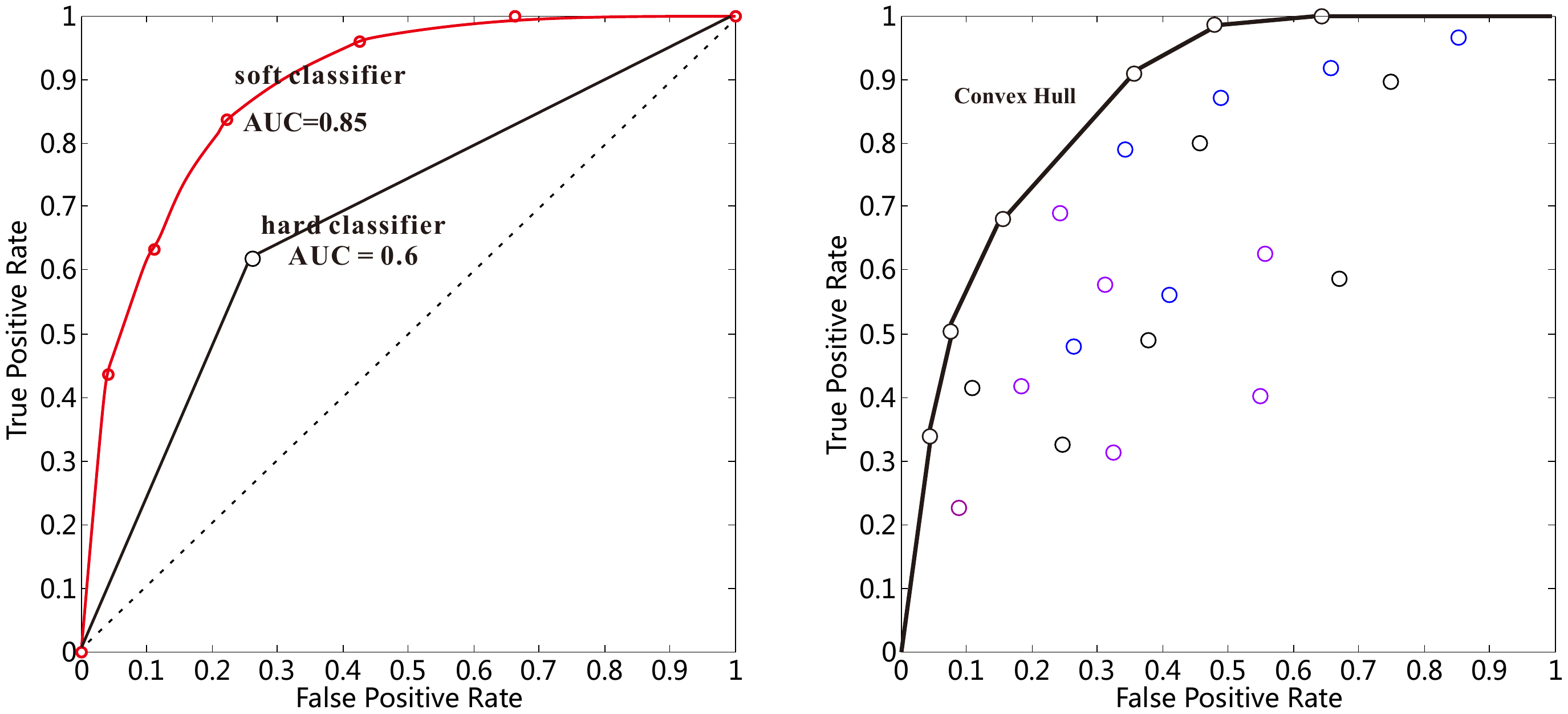}\\
\caption{ROC graph and ROC Convex hull in ROC Space}%
\label{fig:rocspace}%
\end{figure}

Basically, ROC analysis concerns the confusion matrix for the outputs of a classifier, in which we can analysis the performance by measuring different metrics such as accuracy, precise, specificity, sensitivity and some others. ROC graph (Left side of Fig.~\ref{fig:rocspace}) is plotted upon Y axis and X axis respectively taken \emph{tpr} and \emph{fpr}, which are also defined from the confusion matrix. Each classifier can be mapped in the ROC graph by its performance. Essentially, ROCCH is the collection of all potentially optimal classifiers in a given set of classifiers(Right side of Fig.~\ref{fig:rocspace}). Furthermore, a classifier is potentially optimal if and only if it lies on the convex hull of the set of points in ROC space~\cite{fawcett2006introduction}.

\subsection{ROCCH maximization problem and multi-objective optimization problem}
The target of ROOCH maximization problem essentially aims at searching a group of solutions (classifiers)to approximate the upmost line and the leftmost line in ROC space as closely as possible. However, it is conflicting to minimize $fpr$ and maximize $tpr$ simultaneously because if the classifier labels more instances as positives, it will produce less negatives and vice versa. Generally speaking, ROCCH maximization is considered as a multi-objective optimization problem from this perspective and we can describe it as follows:

\begin{eqnarray}%
\label{mop}%
\textnormal{maximize~}F(x) &=& (f_{tpr}(x),f_{1-fpr}(x))\nonumber\\%
\textnormal{subject to~}&&x \in \Omega%
\end{eqnarray}%

In Eq.~\ref{mop}, $x$ is a classifier and $F(x)$ is a vector function for $fpr$ and $tpr$ of the classifier. An important term in MOP is \emph{dominance} which can be defined as: Let $u = (u_{1},\dots,u_{m})$, $v = (v_{1},\dots,v_{m})$ be two vectors, $u$ is said to \emph{dominate} $v$ if $u_{i} \leq v_{i}$ for all $i = 1{\dots}m$, and $u \neq v$, this is noted as $u \prec v$. If $u$ and $v$ can not dominate each other, we say that $u$ and $v$ are \emph{nondominated}. The nondominated set is a set that each item does not dominate any another one. A point $x^{\star}$ is called \emph{Pareto optimal} if there is no $x \in \Omega$ such that $F(x)$ dominates $F(x^{\star})$~\cite{zhang2007moea,WGOEB}. Pareto set (PS) is the collection of all Pareto optimal points. The Pareto front is the set of all the Pareto objective vectors $PF = \{F(x)| x \in PS \}$.

\begin{figure}[htbp]
\centering
\includegraphics[width=0.25\textwidth]{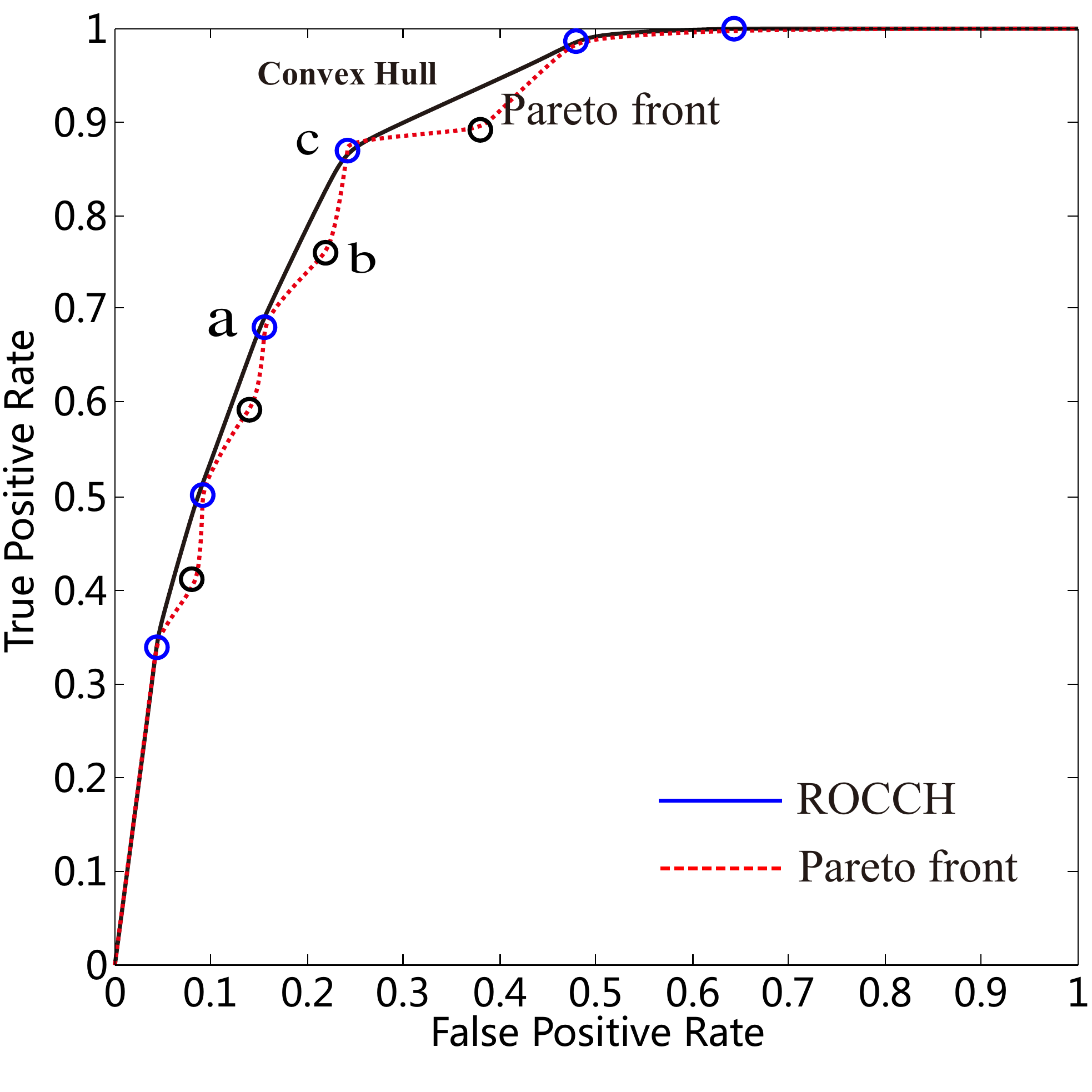}\\
\caption{Pareto front and convex hull}%
\label{fig:ROCCHParetoFront}%
\end{figure}

Most evolutionary multi-objective algorithms involves the pair-wise based dominance to describe the relationship of two solutions. However, we get a special character in ROCCH maximization in ROC space. Fig.~\ref{fig:ROCCHParetoFront} shows the convex hull part and Pareto front for all points. Obviously, convex hull is different from the Pareto front though they were argued that they are similar to each other~\cite{flach2010roc}. For example, points $a,b,c$ in Fig.~\ref{fig:ROCCHParetoFront} are non-dominated set in traditional multi-objective optimization problem, however, the classifier along the line connected by $a$ and $c$ would dominate $b$. That is the special character in ROC maximization problem which makes ROCCH maximization is beyond traditional multi-objective optimization. However, we need to design some new techniques for searching a group of classifiers with maximum ROCCH.

\subsection{Nondominated sort does harm to EMOAs in ROCCH maximization}
\begin{figure}[htbp]
\centering
\includegraphics[width=0.5\textwidth]{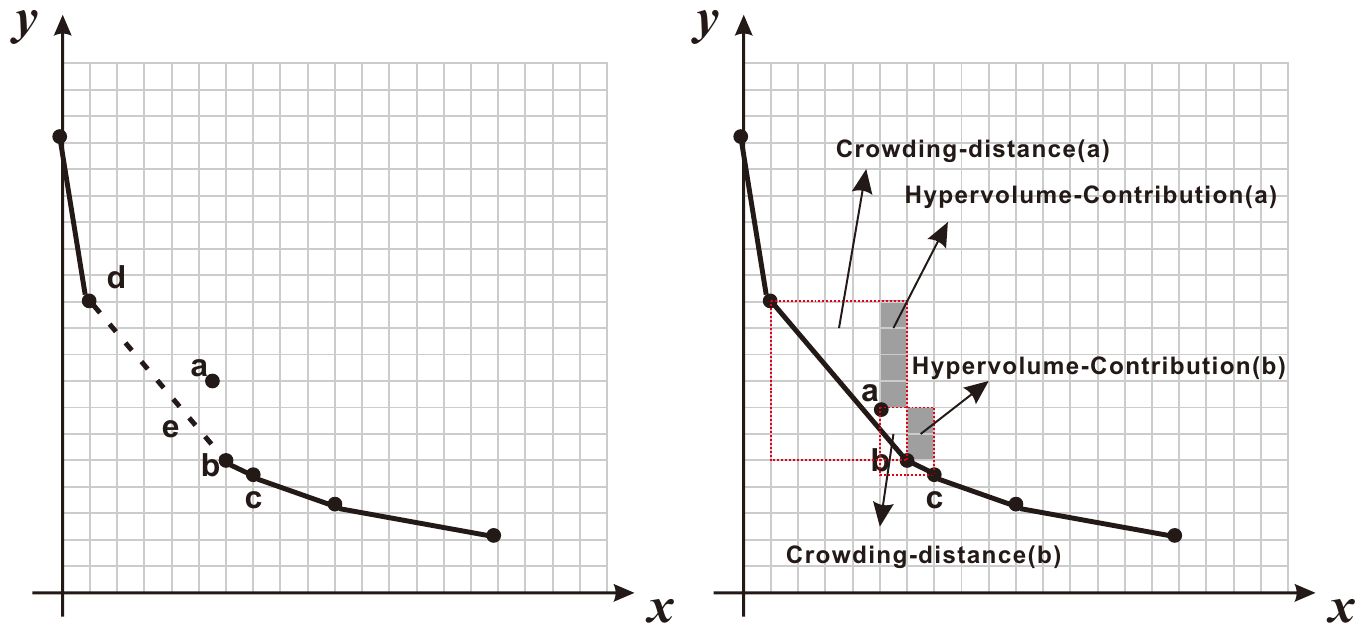}\\
\caption{Nondominated sort keeps the individual which does nothing contribution to ROCCH}%
\label{fig:doharm}%
\end{figure}

The root reason for why we want to get the convex hull rather than Pareto front is that two classifiers will produce any classifiers with their ROC performance which is along the line connected by two point representing for the performance for previous two classifiers in ROC space~\cite{fawcett2004roc}. As shown in left side of Fig.~\ref{fig:doharm}, classifiers with performance at point $d$ and $b$ can be used to construct any virtual classifier with its performance at $e$ along the line connected by $d$ and $b$. That is a special and important character in ROCCH maximization problem.

In the right side of Fig.~\ref{fig:doharm}, all the points are nondominated to each other and belong to the convex hull expect for point $a$. However, if we take crowding-distance selection or hyper-volume contribution based selection to select one individual to be discarded from the population, obviously, point $a$ will be selected to survive rather than point $b$ though point $a$ is not on the convex hull. Actually, there are two phenomenons we need pay attention to, one is the sort strategy and the other is the selection scheme. Besides, suitable sort strategy and selection scheme are should be considered in EMOAs for ROCCH not matter which classifier is involved.

\subsection{The motivation and ideas for new multi-objective algorithms for ROCCH maximization}
We need to think about how to use the special character of ROCCH to make multi-objective optimization techniques more efficient to solve the ROCCH maximization problem. The main techniques for MOP is how to rank the population to select the solutions to survive in next generation. The mostly common rank approach includes two steps, one is sorting the population into several levels indicating the priority level, after that, a selection scheme is used to choose winners from solutions at the same level. In ROCCH maximization problem, firstly, convex hull-based idea will considered into sorting strategy, however, because of the critical concept of convex hull, it would make the diversity decrease fast in the evolutionary process, so we design convex hull-based sorting without redundancy to sort the population. Another idea is to use area-base selection scheme because the target is to maximize the area under the convex hull insteading of hypervolume or crowding-distance. Convex hull-based sorting without redundancy and area-based selection scheme will be descried in detail in Section~\ref{section:chmogp}.
\begin{figure*}[!thbp]
\centering
\includegraphics[width=0.8\textwidth]{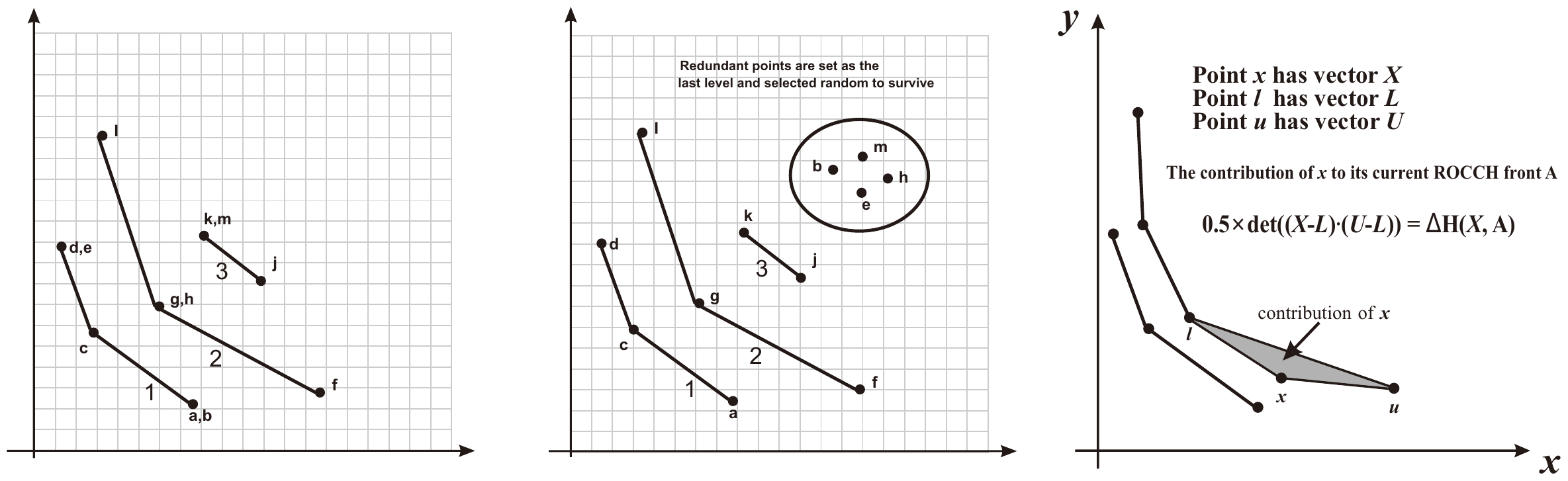}\\
\caption{Convex hull-based sorting with and without redundancy, Area-based contribution to ROCCH}%
\label{fig:chullrankingwithout}%
\end{figure*}

\section{Convex Hull-based Multi-objective Genetic Programming (CH-MOGP)}
\label{section:chmogp}
In this section, we will describe our proposed convex hull-based multi-objective genetic programming to maximize ROCCH. Firstly, convex hull-based sorting without redundancy approach is used to rank the individuals in the union population into several levels which represent different priorities to survive as described in NSGA-II. Secondly, as the target is to maximize the area under the convex hull (AUCH) rather than the hypervolume mentioned in SMS-EMOA, and area-based indicator is designed to calculate the contribution of each individual to AUCH maximization. One major of disadvantage of ($\mu$ + 1) selection strategy was employed in SMS-EMOA and AG-EMOA is that it needs to call fast-nondominated sorting $\mu$ times to select $\mu$ offsprings. In~\cite{igel2007covariance}, an approximate scheme ($\mu$ + $\mu$) is proposed to make the selection process faster, and this idea has been adopted in CH-MOGP.


\subsection{Convex hull-based without redundancy sorting}
\begin{footnotesize}
\begin{algorithm}[!htbp]
\caption{\emph{Convex hull-based-sorting-without-redundancy} ($Q$,$r$)}
\label{algreduce}
\begin{algorithmic}[1]
\REQUIRE $Q  \neq \emptyset$
\STATE   $Q$ is a solution set
\STATE   $r$ is the reference point
\ENSURE  \small{ch-based-sorting-without-redundancy}
\STATE $i$ = 0
\WHILE{$Q \neq \emptyset$}
\STATE   $T$ = $Q \cup \{r\}$
\STATE   $\textbf{F}_i$ = Jarvis-Algorithm($T$)~\cite{jarvis1973identification}
\STATE   $\textbf{F}_i$ = Elimination($\textbf{F}_i$) // Some points in $\textbf{F}_i$ are not interesting and removed
\STATE   $Q = Q - \textbf{F}_i$
\STATE   $i = i+1$
\ENDWHILE
\end{algorithmic}
\end{algorithm}
\end{footnotesize}

First of all, we introduce convex hull-based without redundancy sorting in this subsection. The main idea is too keep the diversity of the population by force, that means, each redundant solutions will be put into an archive to be random selected to survive into the next generation if there is not enough non-redundant solutions to fill the whole population full. Non-redundant solutions with not good performance have chance to be kept by discarding the redundant solutions with good performance to make high diversity, and this could avoid that the solutions at the convex hull being copied a lot at the selection phase in evolutionary multi-objective optimization. As described in Alg.~\ref{algreduce}, the population will be split into redundant part and the other part which is sorted by convex hull-based sorting into several levels and the redundant part is taken as the last level which is the candidates by random selecting.

In Fig.~\ref{fig:chullrankingwithout}, the first and second graphs gives the illustration for convex hull-based sorting with and without redundancy. All the redundant individuals will be discarded into the last level and selected random to the next generation if it is necessary.

\begin{footnotesize}
\begin{algorithm}[!htbp]
\caption{\emph{DeltaArea} ($Q$)}
\label{deltahypervolume}
\begin{algorithmic}[1]
\REQUIRE $Q  \neq \emptyset$
\STATE   $Q$ is a solution set
\ENSURE  \small{DeltaArea}
\STATE $m = sizeof(Q)$
\STATE $\textbf{E}$ is performance of $Q$
\STATE $\textbf{DeltaH}_{1},...,\textbf{DeltaH}_{m} \leftarrow 0$
\IF {$m < 3$}
\STATE Set $\textbf{DeltaH}_{1},...,\textbf{DeltaH}_{m} \leftarrow \infty$
\ELSE
\STATE Set $\textbf{DeltaH}_{1},\textbf{DeltaH}_{m} \leftarrow \infty$
\FOR {$2 \leq i \leq sizeof(Q)-1$}
\STATE $\textbf{DeltaH}_{i}$ = 0.5 $\times$ det(($\textbf{E}_{i}$-$\textbf{E}_{i-1}$) $\circ$ ($\textbf{E}_{i+1}$-$\textbf{E}_{i-1}$))
\ENDFOR
\WHILE {$sizeof(Q) > 2$}
\STATE $r \leftarrow argmin \{\textbf{DeltaH}\}$
\STATE $Q \leftarrow Q \texttt{\char92} \{Q_r \}$
\STATE Update($\textbf{DeltaH}_{r-1}$,$\textbf{DeltaH}_{r+1}$)
\ENDWHILE
\ENDIF
\STATE Return ($\textbf{DeltaH}$)
\end{algorithmic}
\end{algorithm}
\end{footnotesize}

\begin{footnotesize}
\begin{algorithm}[!htbp]
\caption{\emph{Reduce} ($Q$,$N$)}
\label{algreduce}
\begin{algorithmic}[1]
\REQUIRE $Q  \neq \emptyset$
\STATE   $Q$ is a solution set
\STATE   $N$ is the number of solutions will be discarded
\ENSURE  \small{Reduce}
\STATE $F = empty$
\STATE Split $Q$ into two subpopulation $U$ and $R$ // $R$ is the collection of redundant individuals
\IF {$sizeof(R) >= N$}
\STATE   $F \leftarrow $ Random select $N$ solutions from $R$
\STATE   $Q \leftarrow U \cup R \texttt{\char92} F$
\ELSE
\STATE $F \leftarrow R$
\STATE ${\Re_1,...,\Re_v} \leftarrow $$Convex hull$-$based$-$sort$-$without$-$redundancy (Q)$
\FOR   {$i = v ... 1$}
\IF {$sizeof(F)$ + $sizeof(\Re_i) < N$}
\STATE $F \leftarrow F \cup \Re_i$
\STATE $U  = U\texttt{\char92} \Re_i$
\ELSE
\STATE  break
\ENDIF
\ENDFOR
\STATE $T \leftarrow Select$ $(N-sizeof(F))$ solutions with minial $DeltaArea(\Re_i)$
\STATE $F \leftarrow F \cup T$
\STATE $U \leftarrow U \texttt{\char92} T$
\STATE $Q \leftarrow U$
\ENDIF
\STATE Return ($Q$)
\end{algorithmic}
\end{algorithm}
\end{footnotesize}

\subsection{Area-based Selection Scheme}
\begin{equation}%
\Delta area = \frac{det((\textbf{X} - \textbf{L})\circ (\textbf{U} - \textbf{X}))}{2}
\label{equ:area}%
\end{equation}%

In this subsection, we describe our area-based indicator for selection scheme in the new EMOA. The reason for why area-based and not hypervolume-based contribution is adopt is we need to maximize the area under the convex hull. Area-based indicator is more directly and efficiently. In the third graph of Fig.~\ref{fig:chullrankingwithout}, it shows the novel area calculation for two dimensions. The contribution of one point $x$ with its performance vector \textbf{X} to the area is the area of triangle constructed by the point with its predecessor $l$ and successor $u$ with performance vector \textbf{L} and \textbf{U}. Alg.~\ref{deltahypervolume} gives the procedure of calculating of the novel area contribution. Eq.~\ref{equ:area} gives the equation to how to calculate the area contribution of each point to its convex hull front.

\begin{footnotesize}
\begin{algorithm}[!htbp]
\caption{\emph{CH-MOGP} ($Max,N$)}
\label{algchmoea}
\begin{algorithmic}[1]
\REQUIRE $Max > 0, N > 0$
\STATE   $Max$ is the maximum of evaluations
\STATE   $N$ is the population size
\ENSURE  \small{CH-MOGP}
\STATE   $P_{0} = init()$
\STATE   $t = 0$
\STATE   $m = 0$
\WHILE {$m < Max$}
\STATE $Q_{t} = empty$
\FOR {$i = 1:N$}
\STATE $q_{i} \leftarrow$ Operators on $P_t$
\STATE $Q_{t} \leftarrow Q_{t} + q_{i}$
\ENDFOR
\STATE $P_{t+1} \leftarrow Reduce(P_t \cup {Q_{t}})$
\STATE $t \leftarrow t + 1$
\STATE $m \leftarrow m + N$
\ENDWHILE
\end{algorithmic}
\end{algorithm}
\end{footnotesize}
\subsection{CH-MOGP}
Alg.~\ref{algchmoea} describes the CH-MOGP algorithm. The framework is very similar with SMS-EMOA and NSGA-II. However, we employ convex hull-based sorting without redundancy approach to rank the individuals into different levels. ($\mu + \mu$) scheme is adopted into CH-MOGP to maximize the ROC performance. Because the target is to maximize area under the convex hull, area-based selection is designed insteading of hypervolume-based contribution to keep the survivors with high area-based contribution.

In Alg.~\ref{algchmoea}, first of all, the population size and the maximum of evaluations are given. Initial population is constructed by a group of solutions represented by genetic decision trees~\cite{jin2000fgp} using ramped-half-and-half method~\cite{poli2008field}. To generate the offsprings, two operators are employed and described in detail in~\cite{wang2011memetic}. The selection part of CH-MOGP are operated by two schemes like other EMOAs, one is how to sort the population into different levels and the other is how to rank the solutions at the same level. Convex hull-based without redundancy sorting and area-based selection scheme play the main role to the selection part of CH-MOGP. To reduce the time of calling sorting approach, we also take ($\mu$ + $\mu$) scheme not ($\mu$ + 1) in SMS-EMOA.

\section{Experimental Studies}
\label{section:experiment}
\subsection{Data Set}
Nineteen data sets are selected from the UCI repository~\cite{WP27} and described in Table~\ref{DataSets}. Actually, we choose another three large-scaled data sets described in the last row of Table~\ref{DataSets} to make more solid results. In this paper, we focus on binary classification problems, so all the data sets are two-class problems. Balanced and imbalanced benchmark data sets are carefully selected. The scale in terms of the number of instances of these data sets ranges from hundreds to thousands.
\begin{table}[htbp]
\caption{Algorithms Involved}
\label{algorithms}
\begin{center}
\resizebox{0.5\textwidth}{!}{
\begin{tabular}{rccc}
\toprule
Name & Sorting & Selection & Scheme\\
\midrule
\emph{CH-MOGP}   &CH-No-Redundancy& Area             & $\mu + \mu$\\
\emph{RCHH-EMOA} &CH-No-Redundancy& Area             & $\mu + 1$\\
\emph{CH-EMOA}   &Convex Hull     & Hypervolume      & $\mu + 1$\\
\emph{CHCrowding}&CH-No-Redundancy& Crowding-distance& $\mu + \mu$ \\
\emph{CHH-MOGP}  &Convex Hull     & Area             & $\mu + 1$ \\
\emph{NSGA-II}   &Non-dominated   & Crowding-distance& $\mu + \mu$ \\
\emph{SMS-EMOA}  &Non-dominated   & Hypervolume      & $\mu + 1$ \\
\emph{MOEA/D}    &Fitness         & Fitness          & -\\
\bottomrule
\end{tabular}
}
\end{center}%
\end{table}%
\begin{table*}
\caption{Nineteen UCI Data Sets}
\label{DataSets}
\begin{center}
\resizebox{0.7\textwidth}{!}{
\begin{tabular}{rllrllrll}
\toprule
   \multirow{2}{*}{Data Set} &  No. of & Class  &\multirow{2}{*}{Data Set} & No.of & Class &\multirow{2}{*}{Data Set} & No.of & Class \\ & features & Distribution  & & features & Distribution  &  & features & Distribution\\
\midrule
\emph{australian} &14 &383:307    &\emph{house-votes} &16 &168:267  &\emph{pima}  &8   &268:500  \\
\emph{bcw }&9 &458:241    &\emph{ionosphere}  &34 &225:126    &\emph{sonar} &60  &97:111   \\
\emph{crx}  &15 &307:383    &\emph{kr-vs-kp}  &36 &1669:1527    & \emph{monks-3}  &6  &228:204 \\
\emph{transfusion} &4   &178:570 &\emph{mammographic}  &5  &445:516  &\emph{spect}  &22   &212:55   \\
\emph{german} &24 &700:300    &\emph{monks-1} &6  &216:216  &  \emph{parkinsons} &22 &147:48 \\
\emph{wdbc}  &30  &212:357   &\emph{monks-2} &6  &290:142  &\emph{tic-tac-toe} &9  &626:332   \\
 \emph{bands}  &36 &228:312    &  &\\
\bottomrule
\end{tabular}
}
\end{center}%
\end{table*}%

\begin{table*}[htbp]
\caption{Evaluation Times for each algorithm on 19 UCI Data Sets}
\label{EtimesDataSets}
\begin{center}
\resizebox{0.7\textwidth}{!}{
\begin{tabular}{rlrlrlrl}
\toprule
   \multirow{2}{*}{Data Set} &  No. of  &\multirow{2}{*}{Data Set} & No.of &\multirow{2}{*}{Data Set} & No.of &\multirow{2}{*}{Data Set} & No.of \\
   & Evaluations & &  Evaluations  &   & Evaluations&   & Evaluations\\
\midrule
\emph{australian} & 100000 &	\emph{bands} & 150000 &	\emph{bcw} & 50000 &	\emph{crx} & 50000 	\\
\emph{german} & 200000 &	\emph{house-votes} & 30000 &	\emph{ionosphere} & 80000 &	 \emph{kr-vs-kp} & 200000 	\\
\emph{mammographic} & 60000 &	\emph{monks-1} & 200000 &	\emph{monks-2} & 1000000 &	 \emph{monks-3} & 40000 	\\
\emph{parkinsons} & 30000 &	\emph{pima} & 80000 &	\emph{sonar} & 30000 &	\emph{spect} & 40000 	\\
\emph{tic-tac-toe} & 300000 &	\emph{transfusion} & 22000 &	\emph{wdbc} & 30000 &	 \emph{adult} & 10000 	\\
\emph{magic04} & 10000 & 	\emph{skin} & 10000 & &&&\\	
\bottomrule
\end{tabular}
}
\end{center}%
\end{table*}%

\begin{table*}[!htbp]
\caption{Parameters for 8 algorithms}
\begin{center}
\resizebox{0.8\textwidth}{!}{
\begin{tabular}{rlrl}
\toprule
&Objective& Maximize Convex hull in ROC & \\
\midrule
Terminals of GP & \{0,1\} with 1 representing "Positive"; &  Function set of GP & If-then-else , and,\\
& 0 representing "Negative" &&or, not, $>$, $<$ , =. \\
\midrule
Data sets & 22 UCI data sets &
Algorithms & 8 algorithms in Table~\ref{algorithms}\\
\midrule
Crossover rate & 0.9 &
Mutation rate & 0.1 \\
\midrule
Shifting rate & 0.1 &
Splitting rate & 0.1 \\
\midrule
Parameters for GP & P(Population size) = 20; & Termination criterion& Maximum of G of\\
& G (Maximum Evaluation Times) = M & &evaluation time has been reached\\
& Number of Runs : &&\\
& 5 fold cross-validation 20 times& &\\
\midrule
Selection strategy & Tournament selection, Size = 4 &
Max depth of & 3/17\\
&& initial/inprocess individual program & \\
\bottomrule
\end{tabular}%
}
\end{center}%
\label{parametertbl}%
\end{table*}%

\subsection{Algorithms Involved}
To evoluate the performance of two strategies proposed in this paper, Table.~\ref{algorithms} describes the algorithms involved to make rigorous and sufficient experimental comparisons. Generally speaking, this experiment is designed by considering three section of the EMOA, the first one is the strategy in sorting part including convex hull-based with and without redundancy sorting and non-dominated sorting(however, MOEA/D is decomposition based MOEA with different framework), the second one is the indicator for selection schemes including area-based, hyperovlume-based and crowding-distance-based, the last one is related with ($\mu$ + $\mu$) and ($\mu$ + 1) for different EMOAs.

\subsection{Evaluation and Configuration}
\textbf{Evaluation}: To evaluate the generalization performances of different classifiers produced by different algorithms, cross-validation is employed.  We apply each algorithm on each 22 data sets with five-folds cross-validation for 20 times. Because we want to emphasize that our CH-MOGP could be better with less evaluation times, so we run each compared algorithms with large enough evaluation times to make them converge. Table.~\ref{EtimesDataSets} gives the details for algorithms on each data set.

\textbf{Configuration}: We take the representation called GDT~\cite{jin2000fgp} as the individual in all multi-objective evolutionary algorithms. For binary classification problems, 0 and 1 (standing for negative and positive) are selected as the
terminals of GP. Every classifier (individual) is constructed as $if$-$then$-$else$ tree which involves $and$, $or$, $not$, $>$,$<$ and $=$ as operator symbols. Most offspring individuals are obtained by the crossover operator with probability 0.9.  We also employed the shifting, and splitting operators described in~\cite{wang8using} with probability 0.1.  Tournament selection is adopted as the selection strategy and the tournament size is set to 4.  To avoid overfitting, the maximum depth of each individual tree is limited to 17.
\begin{figure*}[htbp]
\centering
\includegraphics[width=\textwidth]{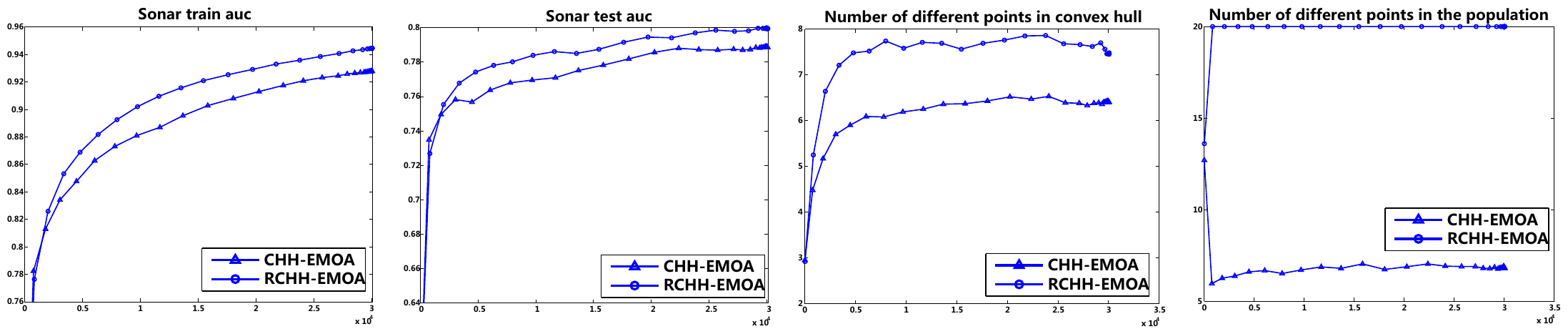}\\
\caption{The diversity in convex hull-based sorting with and without redundancy effects the performance of the results}%
\label{fig:diversity}%
\end{figure*}

\subsection{Results and Analysis}
Fig.~\ref{fig:test}, Fig.~\ref{fig:test2}, Table.~\ref{averagestd} and Table.~\ref{Wilcoxon} show the performance of CH-EMOA compared with other EMOAs in 22 data sets. Generally speaking, CH-EMOA outperforms better not only at the AUCH but also the cost time.

In this subsection, we want to answer the questions as follows:
\begin{enumerate}
\item Why convex hull-based sorting without redundancy is better than traditional convex hull-based sorting?
\item Is convex hull-based sorting without redundancy is better than non-dominated sorting approach in ROCCH maximization problem?
\item Is area-based selection scheme is comparable with or better than crowding-distance or hypervolume based selection?
\item Is CH-MOGP is better than NSGA-II, SMS-EMOA and MOEA/D for ROCCH maximization?
\item Does CH-MOGP show some advantages to traditional machine learning algorithms?
\end{enumerate}

To evaluate the ideas we have proposed, we use 19 data sets in Table.~\ref{DataSets} with algorithms described in Table.~\ref{algorithms}.
\begin{table*}[htbp]
\caption{Performance of four different frameworks of MOGP on UCI data sets, mean and standard
deviation, multiplied by 100, are given in this table}
\label{averagestd}
\begin{center}
\resizebox{\textwidth}{!}{
\begin{tabular}{rcccc|rcccc}
\toprule
&CH-MOGP&SMS-EMOA&NSGA-II&MOEA/D& &CH-MOGP&SMS-EMOA&NSGA-II&MOEA/D\\
\midrule
\emph{australian} &	91.49 $\pm$ 2.72  &  91.67 $\pm$ 2.48  &  91.16 $\pm$ 2.41  &  90.29 $\pm$ 2.75&
\emph{bands} &	77.00 $\pm$ 4.05  &  76.38 $\pm$ 4.09  &  75.54 $\pm$ 3.56  &  71.85 $\pm$ 3.82\\
\emph{bcw} &	97.94 $\pm$ 1.20  &  97.73 $\pm$ 1.56  &  97.84 $\pm$ 1.41  &  97.48 $\pm$ 1.48&
\emph{crx} &	91.30 $\pm$ 2.45  &  91.16 $\pm$ 2.33  &  91.14 $\pm$ 2.36  &  89.88 $\pm$ 2.51\\
\emph{german} &	73.10 $\pm$ 3.24  &  73.32 $\pm$ 3.33  &  72.39 $\pm$ 3.07  &  71.45 $\pm$ 2.85&
\emph{house-votes} &	97.94 $\pm$ 1.56  &  97.69 $\pm$ 1.59  &  97.74 $\pm$ 1.71  &  97.15 $\pm$ 1.75\\
\emph{ionosphere} &	91.07 $\pm$ 4.95  &  90.51 $\pm$ 4.52  &  90.45 $\pm$ 4.53  &  89.89 $\pm$ 4.83&
\emph{kr-vs-kp} &	98.40 $\pm$ 0.89  &  98.63 $\pm$ 0.75  &  98.39 $\pm$ 0.79  &  96.67 $\pm$ 1.43\\
\emph{mammographic} &	89.75 $\pm$ 2.01  &  89.48 $\pm$ 1.94  &  89.41 $\pm$ 1.87  &  87.50 $\pm$ 2.23&
\emph{monks-1} &	99.70 $\pm$ 1.68  &  97.62 $\pm$ 3.71  &  99.62 $\pm$ 1.35  &  96.51 $\pm$ 5.69\\
\emph{monks-2} &	91.05 $\pm$ 8.00  &  89.28 $\pm$ 5.58  &  90.53 $\pm$ 5.19  &  73.26 $\pm$ 9.14&
\emph{monks-3} &	99.81 $\pm$ 0.43  &  99.74 $\pm$ 0.45  &  99.45 $\pm$ 2.87  &  99.07 $\pm$ 0.88\\
\emph{parkinsons} &	86.79 $\pm$ 6.86  &  85.11 $\pm$ 6.68  &  84.90 $\pm$ 7.54  &  83.94 $\pm$ 6.72&
\emph{pima} &	80.08 $\pm$ 3.38  &  79.85 $\pm$ 3.38  &  79.29 $\pm$ 3.70  &  76.93 $\pm$ 3.10\\
\emph{sonar} &	79.42 $\pm$ 5.87  &  78.04 $\pm$ 5.91  &  77.79 $\pm$ 7.34  &  75.75 $\pm$ 5.66&
\emph{spect} &	77.38 $\pm$ 7.36  &  76.27 $\pm$ 7.14  &  76.91 $\pm$ 8.46  &  74.88 $\pm$ 6.43\\
\emph{tic-tac-toe} &	83.40 $\pm$ 10.4  &  79.56 $\pm$ 11.1  &  79.07 $\pm$ 13.4  &  70.85 $\pm$ 10.4&
\emph{transfusion} &	71.62 $\pm$ 4.62  &  71.48 $\pm$ 4.47  &  71.49 $\pm$ 4.84  &  68.77 $\pm$ 4.63\\
\emph{wdbc} &	96.78 $\pm$ 1.92  &  96.49 $\pm$ 2.25  &  96.70 $\pm$ 2.11  &  95.90 $\pm$ 2.19& \\
\bottomrule
\end{tabular}
}
\end{center}%
\end{table*}%

\begin{table*}[htbp]
\caption{Performance of four different frameworks of MOGP on three big data sets, mean and standard
deviation, multiplied by 100, are given in this table}
\label{Laveragestd}
\begin{center}
\resizebox{\textwidth}{!}{
\begin{tabular}{rcccc|rcccc}
\toprule
&CH-MOGP&SMS-EMOA&NSGA-II&MOEA/D& &CH-MOGP&SMS-EMOA&NSGA-II&MOEA/D\\
\midrule
\emph{adult} &	84.58 $\pm$ 1.40  &  82.53 $\pm$ 2.15  &  84.01 $\pm$ 1.38  &  77.04 $\pm$ 2.54&
\emph{magic04} &	83.02 $\pm$ 1.04  &  81.76 $\pm$ 1.57  &  82.01 $\pm$ 1.19  &  76.39 $\pm$ 3.07\\
\emph{skin} &	97.10 $\pm$ 1.11  &  95.46 $\pm$ 1.85  &  96.57 $\pm$ 1.25  &  93.20 $\pm$ 2.37\\
\bottomrule
\end{tabular}
}
\end{center}%
\end{table*}%

\subsubsection{Question 1}
As we argued above, because of the greedy sort of convex hull-based sorting, the diversity will decrease fast as the generation or evaluation times. Fig.~\ref{fig:diversity} shows the performance of CHH-EMOA and RCHH-EMOA which has been described in Table.~\ref{algorithms}. The only different between these two algorithms is the sorting scheme, CHH-EMOA adopts traditional convex hull-based sorting and RCHH-EMOA employs the convex hull-based sorting without redundancy approach. The third and fourth graph in Fig.~\ref{fig:diversity} give the number of different individuals in the convex hull and in the whole population which are simply indicated as the measurement for the diversity. Obviously, RCHH-EMOA with larger diversity performs better than CHH-EMOA is the first and second graph in Fig.~\ref{fig:diversity} which describe the AUCH performance in traning and test data set (Here, we take data set "Sonar" as an example). However, we also give the Wilcoxon-Sum-Rank-Test results (Which is with a
conﬁdence level of 0.95) of RCHH-EMOA and CHH-EMOA for 19 data sets in Table.~\ref{table:Wilcoxon_C1}. Generally speaking, RCHH-EMOA with convex hull-based sorting without redundancy is better than CHH-EMOA.

\begin{table}[htbp]
\caption{Wilcoxon SUM-RANK Test on 19 UCI Data Sets: The table shows the wilcoxon sum test results between RCHH-EMOA and CHH-EMOA on 19 UCI Data sets at different evaluation times. Each $x$-$y$-$z$ in following table means RCH-EMOA wins $x$ times, losses $z$ times and draws $y$ times. Ratio means the ratio of total evaluation times}
\label{table:Wilcoxon_C1}
\begin{center}
\resizebox{0.5\textwidth}{!}{
\begin{tabular}{rlllllll}
\toprule
Ratio & $\frac{1}{15}$ & $\frac{1}{10}$&$\frac{1}{4}$ & $\frac{1}{3}$&$\frac{1}{2}$ & $\frac{2}{3}$&$1$ \\
\midrule
\emph{CHH-EMOA}& 4-15-0 & 5-14-0 & 5-14-0 & 6-13-0 & 6-13-0 & 6-13-0 & 4-15-0\\
\bottomrule
\end{tabular}
}
\end{center}%
\end{table}%

\subsubsection{Question 2}
Algorithm CHCrowding and NSGA-II are involved into answering question 2. As described in Table.~\ref{algorithms}, CHCrowding and NSGA-II employ crowding-distance as the strategy into selection scheme, however, they adopt different sorting approach. Convex hull-based sorting without redundancy is employed into CHCrowding and NSGA-II takes fast nondominated sorting, which is the only difference between them. Table.~\ref{table:Wilcoxon_C2} shows the Wilcoxon-Sum-Rank-Test results (Which is with a
conﬁdence level of 0.95) for them. Obviously, CHCrowding losses none to NSGA-II and wins sometimes.

\begin{table}[htbp]
\caption{Wilcoxon SUM-RANK Test on 19 UCI Data Sets: The table shows the wilcoxon sum test results between CHCrowding and NSGA-II on 19 UCI Data sets at different evaluation times. Each $x$-$y$-$z$ in following table means CHCrowding wins $x$ times, losses $z$ times and draws $y$ times. Ratio means the ratio of total evaluation times}
\label{table:Wilcoxon_C2}
\begin{center}
\resizebox{0.5\textwidth}{!}{
\begin{tabular}{rlllllll}
\toprule
Ratio & $\frac{1}{15}$ & $\frac{1}{10}$&$\frac{1}{4}$ & $\frac{1}{3}$&$\frac{1}{2}$ & $\frac{2}{3}$&$1$ \\
\midrule
\emph{NSGA-II}& 3-16-0 & 2-17-0 & 2-17-0 & 3-16-0 & 3-16-0 & 3-16-0 & 1-18-0\\
\bottomrule
\end{tabular}
}
\end{center}%
\end{table}%

\subsubsection{Question 3}

For question 3, we takes two comparisons to explain. The first is that CHH-EMOA and CH-EMOA which are the same except for the selection schemes. In other words, CHH-EMOA prefers area-based selection and hypervolume contribution is involved into selection scheme for CH-EMOA. We also gives the Wilcoxon-Sum-Rank-Test results (Which is with a
conﬁdence level of 0.95) for them in Table.~\ref{Wilcoxon_C31}. Obviously, area-based selection works better than hypervolume contribution when they are combined with convex hull-based sorting approach into multi-objective optimization algorithm designs. On the other hand, we employ CHCrowding and CH-MOGP to measure the different performance of area-based selection and crowding-distance selection. However, Table.~\ref{Wilcoxon_C3} shows there is no difference between them in 19 data sets. One reason is that convex hull-based sorting without redundancy plays more important role in the multi-objective algorithms than the selection scheme, however, selection scheme is also needed for the EMOAs. Though area-based and crowding-distance based selection schemes show no difference in above two algorithms, we still choose area-based selection because it is more intuitive for maximizing the ROC performance.

\begin{table}[htbp]
\caption{Wilcoxon SUM-RANK Test on 19 UCI Data Sets: The table shows the wilcoxon sum test results between CHH-EMOA and CH-EMOA on 19 UCI Data sets at different evaluation times. Each $x$-$y$-$z$ in following table means CHH-EMOA wins $x$ times, losses $z$ times and draws $y$ times. Ratio means the ratio of total evaluation times}
\label{Wilcoxon_C31}
\begin{center}
\resizebox{0.5\textwidth}{!}{
\begin{tabular}{rlllllll}
\toprule
Ratio of total evaluations & $\frac{1}{15}$ & $\frac{1}{10}$&$\frac{1}{4}$ & $\frac{1}{3}$&$\frac{1}{2}$ & $\frac{2}{3}$&$1$ \\
\midrule
\emph{CH-EMOA}& 3-16-0 & 4-15-0 & 4-15-0 & 4-15-0 & 4-15-0 & 6-13-0 & 5-14-0\\
\bottomrule
\end{tabular}
}
\end{center}%
\end{table}%

\begin{table}[htbp]
\caption{Wilcoxon SUM Test on 19 UCI Data Sets: The table shows the wilcoxon sum test results between CH-MOGP and CHCrowding on 19 UCI Data sets at different evaluation times. Each $x$-$y$-$z$ in following table means CH-MOGP wins $x$ times, losses $z$ times and draws $y$ times. Ratio means the ratio of total evaluation times}
\label{Wilcoxon_C3}
\begin{center}
\resizebox{0.5\textwidth}{!}{
\begin{tabular}{rlllllll}
\toprule
Ratio & $\frac{1}{15}$ & $\frac{1}{10}$&$\frac{1}{4}$ & $\frac{1}{3}$&$\frac{1}{2}$ & $\frac{2}{3}$&$1$ \\
\midrule
\emph{CHCrowding}& 0-19-0 & 0-19-0 & 0-19-0 & 0-19-0 & 0-19-0 & 0-19-0 & 0-19-0\\
\bottomrule
\end{tabular}
}
\end{center}%
\end{table}%

\subsubsection{Question 4}

\textbf{AUCH analysis:} To answer the question 4, we employ more data set, specially for big data set because we always emphasize that our algorithm will perform better with less evaluation times which means we will save a lot of time for problems with expensive evaluation. Table.~\ref{LDataSets} describes three big data set. Table.~\ref{averagestd} and~\ref{Laveragestd} give the result of 4 different evolutionary multi-objective algorithms involved with GDT for maximizing the area under convex hull in ROC space. Furthermore, Table.~\ref{Wilcoxon} gives the Wilcoxon Sum-Rank Test results (Which is with a
conﬁdence level of 0.95) for them. To compare the performance of all algorithms at each stage of its evolutionary process, we show the results at 1/15, 1/10, 1/4, 1/3, 1/2 and 1 of the whole process. It is very clear that CH-MOGP outperforms among these EMOAs.

\begin{table}[htbp]
\caption{Wilcoxon SUM-Rank Test on 22 UCI Data Sets: The table shows the wilcoxon sum test results between CH-EMOA and other three EMOAs (NSGA-II, SMS-EMOA and MOEA/D) on 22 UCI Data sets at different evaluation times. Each $x$-$y$-$z$ in following table means CH-EMOA wins $x$ times, losses $z$ times and draws $y$ times.Ratio means the ratio of total evaluation times}
\label{Wilcoxon}
\begin{center}
\resizebox{0.5\textwidth}{!}{
\begin{tabular}{rlllllll}
\toprule
Ratio & $\frac{1}{15}$ & $\frac{1}{10}$&$\frac{1}{4}$ & $\frac{1}{3}$&$\frac{1}{2}$ & $\frac{2}{3}$&$1$ \\
\midrule
\emph{NSGA-II }& 4-15-0 & 4-15-0 & 2-17-0 & 4-15-0 & 5-14-0 & 5-14-0 & 4-15-0\\
\emph{SMS-EMOA} & 11-8-0 & 11-8-0 & 6-13-0 & 5-14-0 & 4-15-0 & 4-15-0 & 5-14-0\\
\emph{MOEA/D} & 19-0-0 & 19-0-0 & 19-0-0 & 19-0-0 & 19-0-0 & 19-0-0 & 19-0-0\\ \hline\hline
\emph{NSGA-II }& 0-3-0 & 1-2-0 & 1-2-0 & 1-2-0 & 2-1-0 & 2-1-0 & 2-1-0\\
\emph{SMS-EMOA} & 3-0-0 & 3-0-0 & 3-0-0 & 3-0-0 & 3-0-0 & 3-0-0 & 3-0-0\\
\emph{MOEA/D} & 3-0-0 & 3-0-0 & 3-0-0 & 3-0-0 & 3-0-0 & 3-0-0 & 3-0-0\\
\bottomrule
\end{tabular}
}
\end{center}%
\end{table}%

\begin{table*}
\caption{Three Large-scaled Data Sets}
\label{LDataSets}
\begin{center}
\resizebox{0.7\textwidth}{!}{
\begin{tabular}{rllrllrll}
\toprule
   \multirow{2}{*}{Data Set} &  No. of & Class  &\multirow{2}{*}{Data Set} & No.of & Class &\multirow{2}{*}{Data Set} & No.of & Class \\ & features & Distribution  & & features & Distribution  &  & features & Distribution\\
\midrule
\emph{skin}  &4 &50859:194198& \emph{magic04}   &10   &12332 :6688 &  \emph{adult} &14 &11687 : 37155 \\
\bottomrule
\end{tabular}
}
\end{center}%
\end{table*}%

\begin{table*}[htbp]
\caption{Evaluation Times for each algorithm on 22 UCI Data Sets}
\label{LEtimesDataSets}
\begin{center}
\resizebox{0.7\textwidth}{!}{
\begin{tabular}{rlrlrlrl}
\toprule
   \multirow{2}{*}{Data Set} &  No. of  &\multirow{2}{*}{Data Set} & No.of &\multirow{2}{*}{Data Set} & No.of &\multirow{2}{*}{Data Set} & No.of \\
   & Evaluations & &  Evaluations  &   & Evaluations&   & Evaluations\\
\midrule
\emph{australian} & 100000 &	\emph{bands} & 1500000 &	\emph{bcw} & 18500 &	\emph{crx} & 450000 	\\
\emph{german} & 120000 &	\emph{house-votes} & 24000 &	\emph{ionosphere} & 80000 &	 \emph{kr-vs-kp} & 2000000 	\\
\emph{mammographic} & 80000 &	\emph{monks-1} & 230000 &	\emph{monks-2} & 10000000 &	 \emph{monks-3} & 190000 	\\
\emph{parkinsons} & 42000 &	\emph{pima} & 180000 &	\emph{sonar} & 12000 &	\emph{spect} & 10000 	\\
\emph{tic-tac-toe} & 3000000 &	\emph{transfusion} & 35000 &	\emph{wdbc} & 21000 &	 \emph{adult} & 300000 	\\
\emph{magic04} & 40000 & 	\emph{skin} & 30000 & &&&\\	
\bottomrule
\end{tabular}
}
\end{center}%
\end{table*}%

\begin{table*}[htbp]
\caption{Performance of CH-MOGP and traditional classifiers on UCI data sets, mean and standard
deviation, multiplied by 100, are given in this table}
\label{LCCaveragestd}
\begin{center}
\resizebox{\textwidth}{!}{
\begin{tabular}{rcccc|rcccc}
\toprule
 & CH-MOGP & C4.5 & NB & Pyriel &   &     CH-MOGP & C4.5 & NB & Pyriel\\
\midrule
\emph{australian} & 91.97 $\pm$ 2.53 & 85.52 $\pm$ 4.05 & 89.47 $\pm$ 2.78 & 91.75 $\pm$ 2.36 & \emph{monks-3} & 100.0 $\pm$ 0.00 & 100.0 $\pm$ 0.00 & 95.94 $\pm$ 2.17 & 99.60 $\pm$ 0.27\\
\emph{bands} & 78.50 $\pm$ 3.56 & 74.56 $\pm$ 4.59 & 73.91 $\pm$ 4.68 & 76.07 $\pm$ 4.81 & \emph{parkinsons} & 86.10 $\pm$ 6.66 & 78.91 $\pm$ 9.76 & 85.91 $\pm$ 6.11 & 88.24 $\pm$ 5.83\\
\emph{bcw} & 98.17 $\pm$ 1.06 & 95.05 $\pm$ 2.55 & 98.92 $\pm$ 0.62 & 98.16 $\pm$ 1.09 & \emph{pima} & 80.74 $\pm$ 3.12 & 75.23 $\pm$ 4.93 & 81.40 $\pm$ 3.01 & 79.58 $\pm$ 2.92\\
\emph{crx} & 91.82 $\pm$ 2.27 & 85.51 $\pm$ 3.94 & 87.88 $\pm$ 3.16 & 90.65 $\pm$ 2.77 & \emph{sonar} & 81.44 $\pm$ 5.15 & 73.85 $\pm$ 7.84 & 80.12 $\pm$ 7.03 & 69.92 $\pm$ 8.64\\
\emph{german} & 74.27 $\pm$ 2.79 & 65.36 $\pm$ 4.74 & 78.42 $\pm$ 2.94 & 75.95 $\pm$ 3.25 & \emph{spect} & 78.56 $\pm$ 7.44 & 76.88 $\pm$ 8.91 & 84.09 $\pm$ 6.03 & 83.51 $\pm$ 7.01\\
\emph{house-votes} & 98.23 $\pm$ 1.26 & 96.35 $\pm$ 2.04 & 98.05 $\pm$ 1.04 & 97.80 $\pm$ 1.49 & \emph{tic-tac-toe} & 90.07 $\pm$ 8.88 & 84.91 $\pm$ 13.9 & 61.50 $\pm$ 14.7 & 70.41 $\pm$ 12.5\\
\emph{ionosphere} & 92.42 $\pm$ 3.66 & 88.20 $\pm$ 5.65 & 93.57 $\pm$ 3.18 & 93.68 $\pm$ 4.23 & \emph{transfusion} & 72.19 $\pm$ 4.89 & 71.08 $\pm$ 5.08 & 70.93 $\pm$ 4.94 & 70.87 $\pm$ 5.39\\
\emph{kr-vs-kp} & 99.40 $\pm$ 0.26 & 99.71 $\pm$ 0.23 & 93.21 $\pm$ 1.00 & 98.26 $\pm$ 0.44 & \emph{wdbc} & 97.32 $\pm$ 1.40 & 92.74 $\pm$ 3.16 & 98.14 $\pm$ 1.33 & 96.58 $\pm$ 1.94\\
\emph{mammographic} & 90.20 $\pm$ 1.76 & 87.66 $\pm$ 2.21 & 89.77 $\pm$ 1.96 & 89.70 $\pm$ 2.02 & \emph{adult} & 88.97 $\pm$ 0.37 & 88.89 $\pm$ 0.53 & 85.27 $\pm$ 0.37 & 90.37 $\pm$ 0.25\\
\emph{monks-1} & 100.0 $\pm$ 0.00 & 77.13 $\pm$ 6.90 & 73.18 $\pm$ 4.58 & 70.93 $\pm$ 5.59 & \emph{magic04} & 87.16 $\pm$ 0.74 & 86.76 $\pm$ 0.83 & 75.70 $\pm$ 0.74 & 85.37 $\pm$ 0.76\\
\emph{monks-2} & 95.68 $\pm$ 4.61 & 94.17 $\pm$ 5.93 & 52.38 $\pm$ 7.04 & 51.25 $\pm$ 6.16 & \emph{skin} & 99.49 $\pm$ 0.11 & 99.93 $\pm$ 0.02 & 94.17 $\pm$ 0.07 & 98.15 $\pm$ 0.08\\
\bottomrule
\end{tabular}
}
\end{center}%
\end{table*}%

\begin{figure*}[htbp]
\centering
\includegraphics[width=\textwidth]{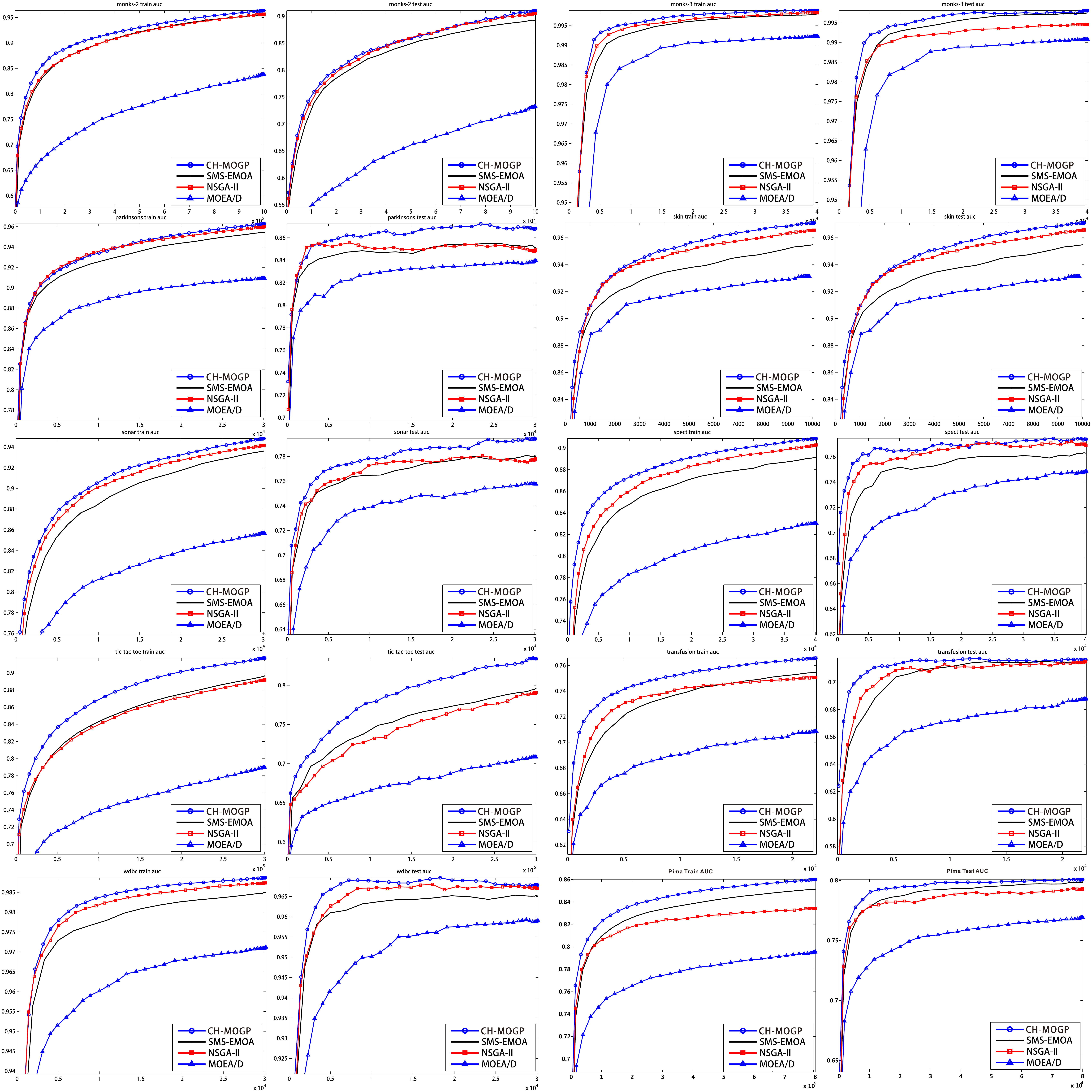}\\
\caption{Performance of four different EMOAs on 10 data sets}%
\label{fig:test2}%
\end{figure*}

\begin{figure*}[htbp]
\centering
\includegraphics[width=\textwidth]{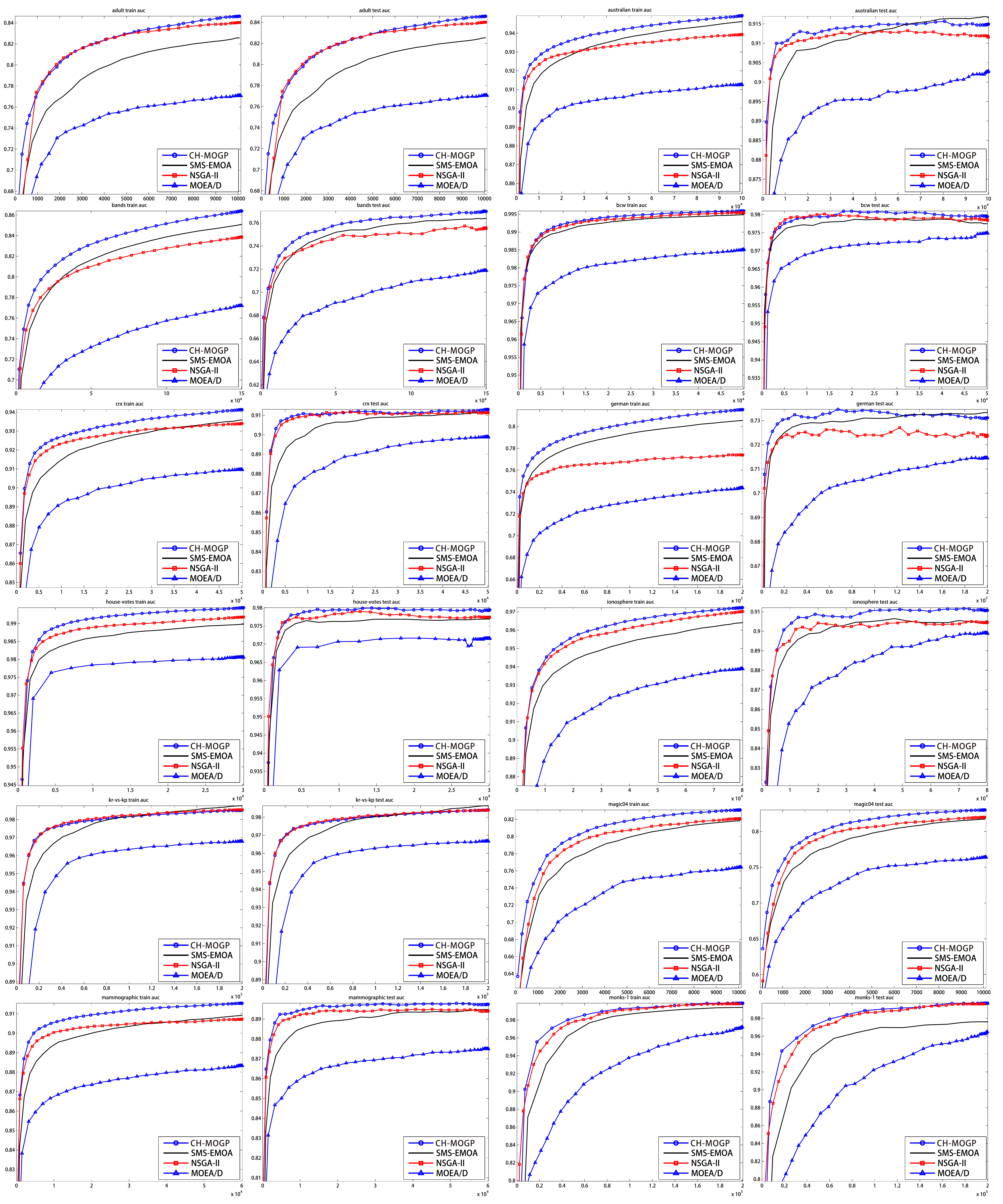}\\
\caption{Performance of four different EMOAs on 12 data sets}%
\label{fig:test}%
\end{figure*}

\begin{table}[htbp]
\caption{Wilcoxon SUM Test on 22 UCI Data Sets: The table shows the wilcoxon SUM-RANK test results between CH-EMOA and other three machine learning algorithms on 22 UCI Data sets at different evaluation times. Each $x$-$y$-$z$ in following table means CH-EMOA wins $x$ times, losses $z$ times and draws $y$ times.}
\label{LWilcoxonT}
\begin{center}
\resizebox{0.4\textwidth}{!}{
\begin{tabular}{rcccc}
\toprule
 & CH-MOGP & C4.5 & NB & PRIE\\
\midrule
\emph{CH-MOGP} &  & 15-5-2 & 11-6-5 & 13-4-5\\
\emph{C4.5} & & & 8-2-12 & 8-1-13\\
\emph{NB} &  &  &  & 6-6-10\\
\emph{Pyriel} &  & & & \\
\bottomrule
\end{tabular}
}
\end{center}%
\end{table}%
\textbf{The Performance and Evaluation Times}:
Fig.~\ref{fig:test} and Fig.~\ref{fig:test2} show the performance of CH-MOGP, SMS-EMOA, NSGA-II and MOEA/D on 22 data sets. Actually, we give the convergence of these EMOAs for training and test data sets with 5-fold cross-validation 20 times. Generally speaking, the curves of CH-MOGP are over others on most data sets. In other words, for a given and very limited evaluation times, CH-MOGP can perform better than other EMOAs in the classification task.

\subsubsection{Question 5}
\textbf{AUCH comparison:} In this sub-section, we compare CH-MOGP with C4.5~\cite{quinlan1993c4}, Naive Bayes(NB)~\cite{lewis1998naive} and PRIE~\cite{fawcett2008prie} which are traditional machine learning algorithms for constructing classifiers. To make a fair comparison, we set the population size of CH-MOGP as 100. The reason is that soft classifiers usually output scores/probabilities to its test data sets, and the number of different kinds of scores or probabilities decides the number of performance points in ROC space, however, that number is not a small one. So we choose a general number, 100, as the population size of CH-MOGP. Meanwhile, it needs more evaluations to a larger population size, so Table.~\ref{LEtimesDataSets} gives the evaluation times for CH-MOGP in 22 data sets. Fig.~\ref{LCCaveragestd} shows the results for CH-MOGP, C4.5, NB and PRIE in all data sets, furthermore, Wilcoxon Sum-Rank Test results (Which is with a conﬁdence level of 0.95) are given in Fig.~\ref{LWilcoxonT}.

\textbf{Evaluation Times:}

\begin{table}[htbp]
\caption{Times for CH-MOGP, C4.5, NB and PRIE to construct classifiers to maximize ROCHH}
\label{Ltime}
\begin{center}
\resizebox{0.55\textwidth}{!}{
\begin{tabular}{rccccrcccc}
\toprule
Time(s) & CHMOGP & C4.5 & NB & PRIE & Time(s) & CHMOGP & C4.5 & NB & PRIE\\
\midrule
australian & 116.91 & 0.06 & 0.02 & 4.18&
bands & 2242.5 & 0.04 & 0.03 & 15.85\\
bcw & 28.63 & 0.01 & 0.02 & 0.53&
crx & 653.45 & 0.02 & 0.02 & 2.92\\
german & 234.27 & 0.16 & 0.04 & 4.79&
house-votes & 13.2 & 0.01 & 0.02 & 0.48\\
ionosphere & 59.51 & 0.04 & 0.02 & 5.77&
kr-vs-kp & 12389.37 & 0.27 & 0.22 & 1.58\\
mammographic & 95.75 & 0.01 & 0.02 & 0.87&
monks-1 & 174.67 & 0.01 & 0.02 & 0.29\\
monks-2 & 8558.14 & 0.01 & 0.02 & 0.3&
monks-3 & 83.49 & 0.01 & 0.02 & 0.31\\
parkinsons & 17.48 & 0.01 & 0.02 & 1.62&
pima & 206.04 & 0.02 & 0.02 & 16.46\\
sonar & 129.28 & 0.03 & 0.02 & 31.45&
spect & 89.05 & 0.02 & 0.02 & 0.39\\
tic-tac-toe & 5396.3 & 0.03 & 0.02 & 0.48&
transfusion & 28.98 & 0.01 & 0.02 & 4.34\\
wdbc & 27.39 & 0.04 & 0.03 & 20.86&
adult & 15655.92 & 0.42 & 2.08 & 1771.73\\
magic04 & 7601.82 & 0.28 & 0.57 & 1103.05&
skin & 91856.38 & 15.01 & 3.7 & 70.15\\
\bottomrule
\end{tabular}
}
\end{center}%
\end{table}%
Table.~\ref{Ltime} gives the cost time for CH-MOGP, C4.5, NB and PRIE to construct classifiers to maximize ROCCH. The experiment environment is an 8 core CPU with 2.13GHz and 24GB RAM. Obviously, CH-MOGP consumes much more time than others, because of the metaheuristic character of EAs, GP needs to evaluate many classiﬁers until it converges. On the other hand, NB method calculates an a posteriori probability and the C4.5 adopts uses a greedy method to increase information gain. PRIE employs a greedy strategy to construct classiﬁers (more than one, usually dozens of classiﬁers) to maximize the ROCCH, so it cost a little more time than NB and C4.5, but still much less than CH-MOGP. Actually, how to reduce the evaluation time of CH-MOGP is an important topic.
\section{Conclusions and Future Work}
\label{section:confusion}
In this paper, we propose convex hull-based sorting approach and area-based selection scheme involved into multi-objective genetic programming for maximizing the ROC performance in classification tasks. First, we emphasized that convex hull maximization problems is similar but beyond multi-objective optimization problem, traditional techniques are helpful but needed to improve the solve the this kinds of problem. Insteading of fast non-dominated sorting approach in NSGA-II and SMS-EMOA, convex hull-based sorting is investigated in new algorithm design, however, we found convex hull-based sorting without redundancy was efficient to avoid losing diversity in the search process. Area-based selection scheme with $\mu + \mu$ is also designed for helping to rank the population. The new algorithm- CH-MOGP is also performed on benchmarks and work better than other traditional EMOAs and some other traditional machine learning algorithms. In the future work, there are three topics would be discussed. The first is how to improve CH-MOGP to reduce its time consuming character but keep the comparable performance for ROCCH maximization. The second one is that GP-based classifier could be replaced by other tree-based classifiers or other traditional machine learning classifiers such as SVM, NB, etc.. Different classifier would result better performance for ROCCH maximization. The third topic is convex hull based without redundancy sorting and area-based selection scheme, these two strategies are not only used in classification but also other area such as numerical optimization.

\section*{Acknowledgment}

The authors would like to thank...

\bibliographystyle{elsarticle-num}
\bibliography{refer}
%



%





\end{document}